\documentclass{files/ecai} 
\usepackage{latexsym}
\usepackage{amssymb}
\usepackage{amsmath}
\usepackage{amsthm}
\usepackage{booktabs}
\usepackage{enumitem}
\usepackage{graphicx}
\usepackage{color}
\usepackage{makecell} % Include this in the preamble
\usepackage{booktabs} % For professional-looking tables
\usepackage{listings}
\usepackage{xcolor} % Optional: for better code highlighting
\usepackage{caption} % Optional: for caption formatting with listings
\newcommand{\BibTeX}{B\kern-.05em{\sc i\kern-.025em b}\kern-.08em\TeX}

\begin{document}

%%%%%%%%%%%%%%%%%%%%%%%%%%%%%%%%%%%%%%%%%%%%%%%%%%%%%%%%%%%%%%%%%%%%%%%%

\begin{frontmatter}

%%% Use this command to specify your submission number.
%%% In doubleblind mode, it will be printed on the first page.

\paperid{123} 

%%% Use this command to specify the title of your paper.

\title{Human-in-the-Loop: Quantitative Evaluation of 3D Models Generation by Large Language Models}

%%% Use this combinations of commands to specify all authors of your 
%%% paper. Use \fnms{} and \snm{} to indicate everyone's first names 
%%% and surname. This will help the publisher with indexing the 
%%% proceedings. Please use a reasonable approximation in case your 
%%% name does not neatly split into "first names" and "surname".
%%% Specifying your ORCID digital identifier is optional. 
%%% Use the \thanks{} command to indicate one or more corresponding 
%%% authors and their email address(es). If so desired, you can specify
%%% author contributions using the \footnote{} command.

\author[A]{\fnms{Ahmed R.}~\snm{Sadik}\orcid{0000-0001-8291-2211}\thanks{Corresponding Author. Email: ahmed.sadik@honda-ri.de.}}
\author[B]{\fnms{Mariusz}~\snm{Bujny}\orcid{0000-0003-4058-3784}}

\address[A]{Honda Research Institute Europe - Germany}
\address[B]{NUMETO - Poland}

%%%%%%%%%%%%%%%%%%%%%%%%%%%%%%%%%%%%%%%%%%%%%%%%%%%%%%%%%%%%%%%%%%%%%%%%
%%% Use this environment to include an abstract of your paper.
% Place figure right after authors/affiliations

\begin{abstract}

Large Language Models (LLMs) are increasingly capable of interpreting multimodal inputs to generate complex 3D shapes, yet robust methods to evaluate geometric and structural fidelity remain underdeveloped. This paper introduces a human-in-the-loop framework for the quantitative evaluation of LLM-generated 3D models, supporting applications such as democratization of CAD design, reverse engineering of legacy designs, and rapid prototyping. We propose a comprehensive suite of similarity and complexity metrics—including volumetric accuracy, surface alignment, dimensional fidelity, and topological intricacy—to benchmark generated models against ground-truth CAD references. Using an L-bracket component as a case study, we systematically compare LLM performance across four input modalities: 2D orthographic views, isometric sketches, geometric structure trees, and code-based correction prompts. Our findings demonstrate improved generation fidelity with increased semantic richness, with code-level prompts achieving perfect reconstruction across all metrics. A key contribution of this work is demonstrating that our proposed quantitative evaluation approach enables significantly faster convergence toward the ground truth, especially compared to traditional qualitative methods based solely on visual inspection and human intuition. This work not only advances the understanding of AI-assisted shape synthesis but also provides a scalable methodology to validate and refine generative models for diverse CAD applications.
\end{abstract}
\end{frontmatter}

%%%%%%%%%%%%%%%%%%%%%%%%%%%%%%%%%%%%%%%%%%%%%%%%%%%%%%%%%%%%%%%%%%%%%%%%

\section{Introduction}

\begin{figure}[!h]
    \centering
    \includegraphics[width=\linewidth]{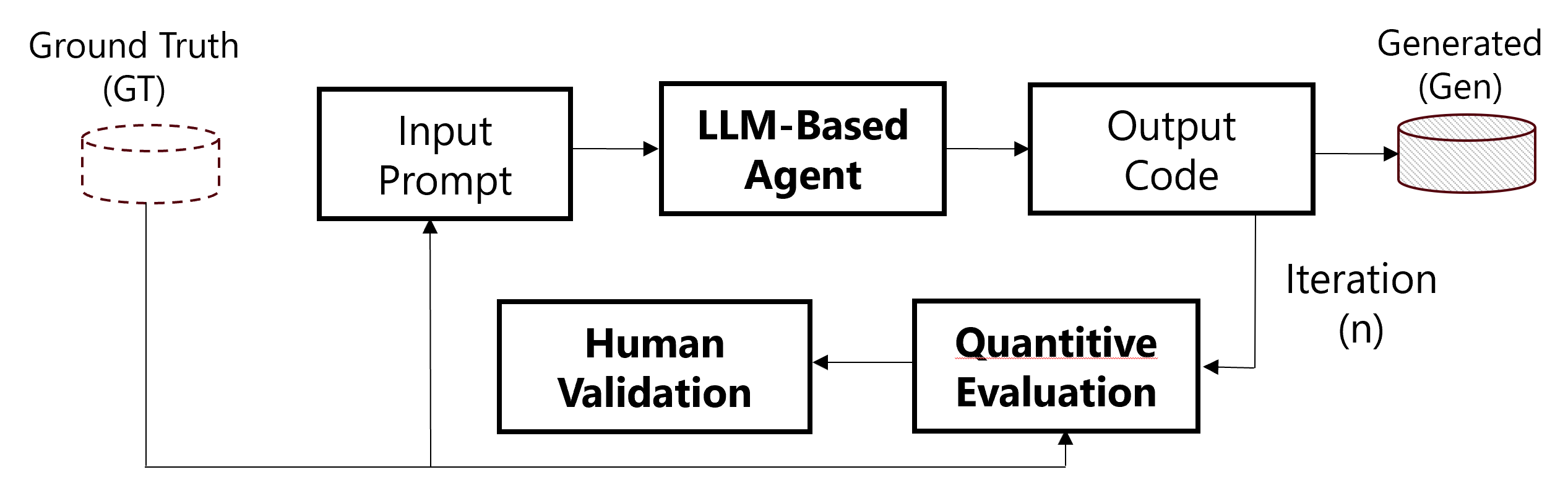}
    \caption{Human-in-the-loop for iterative 3D model generation.}
    \label{fig:system_architecture}
\end{figure}

\vspace{0.4cm}

Large Language Models (LLMs) have shown remarkable progress in processing multimodal information and generating structured outputs, particularly in the realm of 3D geometry synthesis. This capability offers transformative potential for various Computer-Aided Design (CAD) applications, significantly advancing the democratization of design by making sophisticated modeling accessible to non-experts, enabling efficient reverse engineering and modification of legacy designs, and enhancing rapid prototyping and iterative product development. However, despite these promising capabilities, practical adoption of LLM-generated models in critical engineering and manufacturing tasks remains limited due to the absence of rigorous evaluation protocols and structured feedback mechanisms~\cite{yoshimura2024variability, sadik2023analysis}. Without reliable quantitative evaluation, generated models may lack the dimensional accuracy, structural consistency, and geometric integrity necessary for real-world applications.

One critical motivating domain for this research is the conversion and preservation of legacy engineering designs. Industrial and educational archives frequently contain vast collections of 2D orthographic drawings and isometric sketches that require significant manual effort and domain-specific expertise to convert accurately into modern digital CAD formats. Automating this translation process using LLM-based systems not only reduces intensive manual tasks but also facilitates broader access to digital design tools across diverse user groups, including smaller enterprises, educators, and independent designers. Additionally, explicit quantitative feedback enabled by structured evaluation metrics can streamline iterative design refinement, significantly benefiting rapid prototyping, customized product development, and reverse engineering workflows~\cite{paviotiterative}.

To address these opportunities, we propose a human-in-the-loop framework that integrates generation, evaluation, and iterative feedback into a structured pipeline, as shown in Figure~\ref{fig:system_architecture}. This approach leverages multiple input modalities, ranging from conventional 2D drawings and sketches to structured semantic descriptions and symbolic CAD code. Generated models are quantitatively assessed against ground-truth references using comprehensive complexity and similarity metrics, offering explicit and actionable feedback to guide human-driven refinement steps. Our study demonstrates this methodology through a detailed case study involving an L-bracket component, quantitatively evaluating the effectiveness of LLM-based generation from different input formats~\cite{ocker2025idea}. Results highlight the critical role of semantic richness and structural clarity in input prompts, with symbolic CAD instructions achieving the highest fidelity. By systematically quantifying the generated outputs quality, our approach supports robust applications including quality assurance in manufacturing, retrieval and reuse in digital CAD libraries, educational training, and cross-disciplinary cooperative design.

The remainder of this paper is structured as follows: Section~\ref{sec:soa} reviews related work on LLM-assisted design and 3D shape evaluation. Section~\ref{sec:evaluation} describes our proposed evaluation metrics, Section~\ref{sec:case} presents the case study and analysis of results, and Section~\ref{sec:conclusion} concludes by summarizing insights and outlining future research directions.

%%%%%%%%%%%%%%%%%%%%%%%%%%%%%%%%%%%%%%%%%%%%%%%%%%%%%%%%%%%%%%%%%%%%%%%%

\section{State Of the Art}
\label{sec:soa}
This section provides an overview of the literature on LLM-based CAD model generation in engineering design, followed by a discussion of related work on metrics for quantifying complexity and geometric differences between mechanical parts.

\subsection{LLM-assisted Engineering Design}
\label{sec:soa_llm_design}
The popularity of LLMs in recent years, supported by their demonstrated capabilities---especially in tasks related to text processing, such as code generation---has led to increased interest from the research community in applying such models in the domain of mechanical design.

Early works on the topic focused on providing engineers with inspiration for conceptual designs by utilizing the text-to-text capabilities of models like GPT-2 and GPT-3~\cite{zhu_generative_2022, zhu_generative_2023, zhu_biologically_2023}. In these studies, the authors evaluated LLMs beyond simple prompting, exploring few-shot learning and model fine-tuning, also incorporating specialized datasets, such as those for bio-inspired design~\cite{zhu_biologically_2023}. 

Despite the potential value of using LLMs to support engineers in the creative design process through textual descriptions of new concepts, the success of LLMs in code generation expanded opportunities for their use in directly generating geometric models by producing scripts in CAD libraries and mechanical simulation software. In one of the first comprehensive studies evaluating the potential of using LLMs to support humans in design and manufacturing based on code generation, Makatura et al.~\cite{makatura_how_2023} assessed the capabilities of the GPT-4 model in tasks such as generating parametric CAD models from high-level textual descriptions, selecting optimal manufacturing processes and generating manufacturing instructions, and evaluating performance metrics based on design features such as geometry and materials, including Finite Element (FE) simulations. By providing a wide range of application scenarios, this work demonstrated the high potential of LLMs for solving engineering design problems by applying a complete process,  from conceptual design, through appropriate parametrization of geometry, to evaluation of mechanical properties and optimization of designs. 

Following this direction, Picard et al.~\cite{picard_concept_2023} presented an extensive evaluation of vision-language models for engineering design based on the GPT-4V model. The authors addressed a wide range of problems, including conceptual design from both textual and graphical inputs, CAD model generation from technical drawings, material selection, interpretation of numerical simulation results, inspection of designs for specific manufacturing processes, and the use of vision LLMs for engineering education tasks. The broad spectrum of problems related to the application of vision LLMs in engineering design, covered in the article, was further explored in a series of problem-specific papers. Alrashedy et al.~\cite{alrashedy_generating_2024} introduced a framework for creating 3D objects through iterative refinements, using a vision LLM to generate Python code for parametric CAD models. Similarly, Li et al.~\cite{li_llm4cad_2024} used textual and graphical inputs to generate CAD models, relying on commercial LLMs and the CadQuery scripting language. Other works~\cite{yuan_openecad_2024, wu_cad-llm_nodate} explored the concept of model fine-tuning to obtain specialized models capable of generating 2D sketches from partial input drawings and 3D CAD models based on mechanical part views, respectively. Jadhav et al.~\cite{jadhav_large_2024} studied the capability of LLMs for optimizing designs by integrating them with an FE module, which they demonstrated through an iterative optimization of truss structures. Finally, in~\cite{wang_chatgpt_2023, hou_assessing_2024, doris_designqa_2024}, the use of vision LLMs for engineering education and documentation understanding was further explored.

An interesting, separate research direction is the incorporation of LLMs into numerical optimization approaches. Rios et al.~\cite{rios_large_2023} investigated the problem of aerodynamic vehicle optimization using a text-to-3D Shape-E framework. In this approach, text prompts were optimized based on an evolutionary algorithm to generate 3D surface meshes of cars, which were then used in Computational Fluid Dynamics (CFD) simulations to evaluate the fitness of different prompts based on the drag coefficient. This approach was further extended by incorporating a vision LLM to penalize impractical designs~\cite{wong_generative_2024}.

A natural next step in the evolution of LLM-based frameworks for supporting engineering design is the development of Multi-Agent Systems (MAS) composed of models specialized in different tasks within the product development process. This concept was explored in~\cite{ni_mechagents_2024}, where the authors proposed a system of multiple LLM agents collaborating to solve problems in structural mechanics. By enabling collaboration and mutual correction among specialized agents, the authors claim an overall improvement in system performance. Finally, most recent works---specifically the multi-agent architecture of Ocker et al.~\cite{ocker2025idea}, which integrates agents responsible for requirements engineering, CAD modeling, and quality assurance to automatically generate 3D models from sketches and textual descriptions---demonstrate how MAS can effectively coordinate complex engineering workflows, share intermediate knowledge, and deliver expert-level designs with minimal human intervention.

A key element underpinning many of the methods discussed in this section---yet often not given sufficient emphasis---is the use of specialized metrics and methodologies for quantifying geometric differences between the generated and target designs. In the next section, we discuss related works from the engineering domain that inspire the approaches proposed in this work.

\subsection{Complexity and Similarity Metrics}
\label{sec:soa_metrics}

Quantitative evaluation of the geometric properties of 3D objects plays a crucial role across a wide range of fields, including computational geometry, computer graphics, geometric deep learning, and mechanical engineering. As a result, numerous well-established metrics have been developed to assess both the geometric complexity of individual objects and the similarity between different 3D shapes. In this paper, we focus specifically on the metrics relevant to engineering design, with particular emphasis on their application within design processes supported by LLMs.

Although there are no standard, objective metrics for evaluating the complexity of CAD models \cite{amadori_flexible_2012, johnson_investigation_2018}, the topic has been widely studied due to its potential impact on education and research~\cite{summers_mechanical_2010}, as well as on practical applications such as assessing manufacturability~\cite{kerbrat_manufacturability_2010, joshi_quantifying_2010}, cost~\cite{mandolini_analytical_2019, bujny_method_2024}, or reusability within the context of product lifecycle management~\cite{salehi_methodological_2009}. Common metrics of geometric complexity include the number of surfaces, the number of triangles or vertices in an STL model, volume ratio, sphere area ratio, and cube ratio~\cite{joshi_quantifying_2010}. A particularly comprehensive study of various complexity metrics in relation to perceived complexity was conducted by Johnson et al.~\cite{johnson_investigation_2018}, which found that metrics involving the bounding box properties---such as volume ratio—correlate strongly with subjective assessments of complexity made by human evaluators.

In contrast to the complexity metrics, the importance of geometric similarity metrics of mechanical parts started increasing recently, particularly with the advent of data mining~\cite{graening_shape_2014} and machine learning techniques in the context of engineering design. By quantifying similarity---even using hand-engineered features based on material distribution, surface curvature, spectral descriptors~\cite{lara_lopez_comparative_2017}, or point cloud and voxel representations---automatic shape retrieval and concept identification in large datasets of engineering components became possible~\cite{dommaraju2023evaluation}. More advanced use cases of such metrics include their direct incorporation into loss functions in geometric deep learning~\cite{achlioptas_learning_2017, wollstadt_carhoods10k_2022}, as well as their application in similarity-driven 3D object synthesis via topology optimization~\cite{yousaf_similarity_2021, yousaf_similarity-driven_2023}.

In the context of the LLM-based generation of parametric CAD models, many of the works rely predominantly on manual, qualitative evaluation of the resulting models~\cite{makatura_how_2023, picard_concept_2023, li_llm4cad_2024} to assess similarity to the ground truth. Going beyond that, Picard et al.~\cite{picard_concept_2023} introduced a scoring system to evaluate the ability of LLMs to generate CAD models based on the provided engineering drawings. In their approach, points were awarded based on visual inspection of the designs, according to a set of predefined criteria---such as the presence of specific geometric features (e.g., holes) and the correctness of key dimensions. Moving toward more automated evaluation, Li et al.~\cite{li_llm4cad_2024} proposed a method that quantifies design reconstruction quality using the Intersection over Union (IoU) between generated and target 3D CAD models, after alignment via principal axis-based rotation and translation. In a similar spirit, Alrashedy et al.~\cite{alrashedy_generating_2024} introduced the Intersection over Ground Truth (IoGT) metric, which measures the overlap between the bounding boxes of the 3D objects. Finally, approaches based on point cloud representations have also been proposed~\cite{yuan_3d-premise_2024, alrashedy_generating_2024}, in which 3D objects are converted into point clouds, normalized by aligning the generated CAD models with the underlying ground truth objects, and the similarity between them is evaluated using standard metrics such as Euclidean distance between matching nearest points, Chamfer distance, or Hausdorff distance.

Despite the value of the proposed methods, a research gap remains in the systematic evaluation of metric properties, particularly in the context of LLM-based engineering design. In this work, we aim to bridge this gap by analyzing the characteristics of a proposed suite of complexity and similarity metrics for AI-assisted synthesis of CAD models.

%%%%%%%%%%%%%%%%%%%%%%%%%%%%%%%%%%%%%%%%%%%%%%%%%%%%%%%%%%%%%%%%%%%%%%%%
\section{Evaluation Metrics for LLM-Generated 3D Models}
\label{sec:evaluation}

To rigorously evaluate the outputs of LLMs in generating 3D geometry, we propose a dual-metric evaluation framework that considers both the structural complexity of the generated models and their geometric similarity to a known ground truth. This approach enables us to measure not only how well the generated model matches its intended design, but also how detailed, intricate, or challenging the shape is in terms of its structure. By correlating complexity with similarity, we can benchmark complete 3D shapes databases, to better understand the performance limits of different input modalities and generation strategies, and determine whether higher complexity leads to lower fidelity, or if certain input types enable better performance despite increased geometric intricacy.

The evaluation framework consists of two distinct but complementary metric sets:

\begin{itemize}
    \item Structural complexity metrics, which assess the internal richness and topological detail of the 3D model ground truth.
    \item Geometric similarity metrics, which quantify how closely the generated shape reproduces the ground-truth model in terms of dimensions, volume, surface fidelity, and alignment.
\end{itemize}

Together, these metrics help the human to identify limitations in the generation methods related to shape complexity, guiding iterative collaboration with the LLM by highlighting what went wrong in the previous iteration and which aspects of the 3D model need adjustment to better match the ground truth.

\subsection{Structural Complexity Metrics}

Measuring the structural complexity of the 3D model provides insight into the level of detail and design richness produced by the LLM, and the limitation of each agent architecture to generate the 3D model. Furthermore, these metrics help identify whether certain input modalities yield overly simplistic outputs or if the generated shapes capture the fine-grained intricacies of the intended design.

\textbf{Feature Complexity ($C_f$):} Captures the total number of triangular faces in the STL mesh representation:
\begin{equation}
    C_f = F
\end{equation}
where $F$ is the number of mesh faces. This directly reflects the geometric resolution of the shape; more faces indicate more detailed or elaborate features.

\textbf{Surface Complexity ($C_s$):} Reflects the ratio between surface area and volume:
\begin{equation}
    C_s = \frac{A}{V}
\end{equation}
where $A$ is the total surface area and $V$ is the volume enclosed by the model. This highlights thin or contoured designs that may be more difficult to synthesize or fabricate.

\textbf{Topological Complexity ($C_t$):} Measured using the Euler characteristic of the mesh:
\begin{equation}
    C_t = X - E + F
\end{equation}
where $X$, $E$, and $F$ represent the number of vertices, edges (approximated as $1.5 \times F$ for triangular meshes), and faces, respectively. This captures features such as holes or tunnels that increase topological intricacy.

\textbf{Composite Complexity Score ($C$):} A weighted combination of all three components:
\begin{equation}
    C = K_1 \cdot C_f + K_2 \cdot C_s + K_3 \cdot C_t
\end{equation}
where $K_1$, $K_2$, and $K_3$ control the influence of each term. In our experiments, $C_f$ is prioritized to emphasize geometric detail, while $C_s$ and $C_t$ balance the evaluation with surface and topological insights. Please note that the coefficients can also be fine-tuned through regression to better align with the complexity perceived by human designers.

\subsection{Generation Similarity Metrics}

To evaluate how accurately the generated 3D model replicates the intended geometry, we introduce a set of complementary metrics. These capture both global properties (e.g., volume, dimensions) and fine-grained spatial relationships (e.g., alignment, surface deviation). Together, they provide a comprehensive assessment of structural fidelity, which is essential for tasks involving engineering design, manufacturing, or simulation.

\textbf{Dimensional Accuracy ($S_d$):}  
Captures deviation in key dimensions (length, width, height) between the generated and reference models:
\begin{equation}
    S_d = 1 - \frac{|D_g - D_t|}{D_t}
\end{equation}
where $D_g$ and $D_t$ represent a selected dimension (or averaged dimensions) from the generated and ground-truth models. This metric is critical in engineering applications where precise sizing affects part interoperability and fit.

\textbf{Volumetric Similarity ($S_v$):}  
Measures the agreement in overall volume:
\begin{equation}
    S_v = 1 - \frac{|V_g - V_t|}{V_t}
\end{equation}
where $V_g$ and $V_t$ are the volumes of the generated and target models, respectively. Volume offers a global geometric check and helps identify under- or over-generated material regions.

\textbf{Surface Similarity ($S_a$):}  
Evaluates how closely the surface area of the generated model matches the reference:
\begin{equation}
    S_a = 1 - \frac{|A_g - A_t|}{A_t}
\end{equation}
where $A_g$ and $A_t$ are the surface areas. This reflects the fidelity of external contours, surface detail, and small features that may not significantly impact volume but are important visually or functionally.

\textbf{Hausdorff Distance ($H_d$):}  
Quantifies the maximum surface deviation between two point sets:
\begin{equation}
    H_d = \max \left( \sup_{p \in G} \inf_{q \in T} d(p,q), \sup_{q \in T} \inf_{p \in G} d(p,q) \right)
\end{equation}
where $G$ and $T$ are point clouds sampled from the generated and target surfaces. Unlike global metrics, Hausdorff Distance captures \textbf{worst-case local errors}, making it useful for detecting sharp misalignments, missing features, or small deformations.

\textbf{Principal Component Analysis (PCA) Alignment Score ($S_p$):}  
Assesses high-level orientation similarity by comparing the principal axes of the two shapes:
\begin{equation}
    S_p = 1 - \frac{\| C_g - C_t \|}{\| C_t \|}
\end{equation}
where $C_g$ and $C_t$ are the principal component vectors derived from PCA on each shape’s point cloud. This metric offers a quick estimation of shape orientation and is especially helpful as a coarse alignment check or pre-processing step for finer metrics like ICP.

\textbf{Iterative Closest Point (ICP) Alignment Score ($S_i$):}  
Measures point-wise alignment quality after optimal rigid registration: where $S_i \in [0, 1]$, with higher values indicating better spatial alignment.  
ICP iteratively minimizes the distance between the closest points of two clouds, refining alignment through rotation and translation. It provides a \textbf{precise post-registration similarity measure}, reflecting geometric congruence even after partial or rotated mismatches.

\textbf{Final Similarity Score ($S_f$):}  
An aggregated score combining the most informative metrics:
\begin{equation}
    S_f = K_1 \cdot S_v + K_2 \cdot S_a + K_3 \cdot S_d + K_4 \cdot S_p + K_5 \cdot S_i
\end{equation}
The weights $K_1$ to $K_5$ can be tuned based on task requirements, e.g., by adjusting them manually or through regression, to match similarity perceived by humans. In this work, we prioritize volume and surface similarity due to their importance in maintaining CAD model integrity and manufacturability. This composite metric provides a holistic measure of fidelity across scales and geometric attributes.

%%%%%%%%%%%%%%%%%%%%%%%%%%%%%%%%%%%%%%%%%%%%%%%%%%%%%%%%%%%%%%%%%%%%%%%%

\section{Case Study and Experimental Analysis}
\label{sec:case}

To investigate the quantitative capabilities of LLMs in generating accurate and structurally faithful 3D models, we present a case study focused on evaluating the geometry synthesis process using an L-bracket component. This scenario is selected not only for its moderate geometric complexity but also for its diverse structural features—such as through-holes, cutouts, and sharp dimensional constraints—that make it particularly well-suited for rigorous metric-based benchmarking. The L-bracket serves as an ideal testbed for assessing fidelity using our proposed evaluation framework, which incorporates structural complexity and geometric similarity metrics to quantify the accuracy of LLM-generated outputs across multiple input modalities.

\subsection{Input Modalities}

\begin{figure}[!ht]
    \centering
    \includegraphics[width=0.6\linewidth]{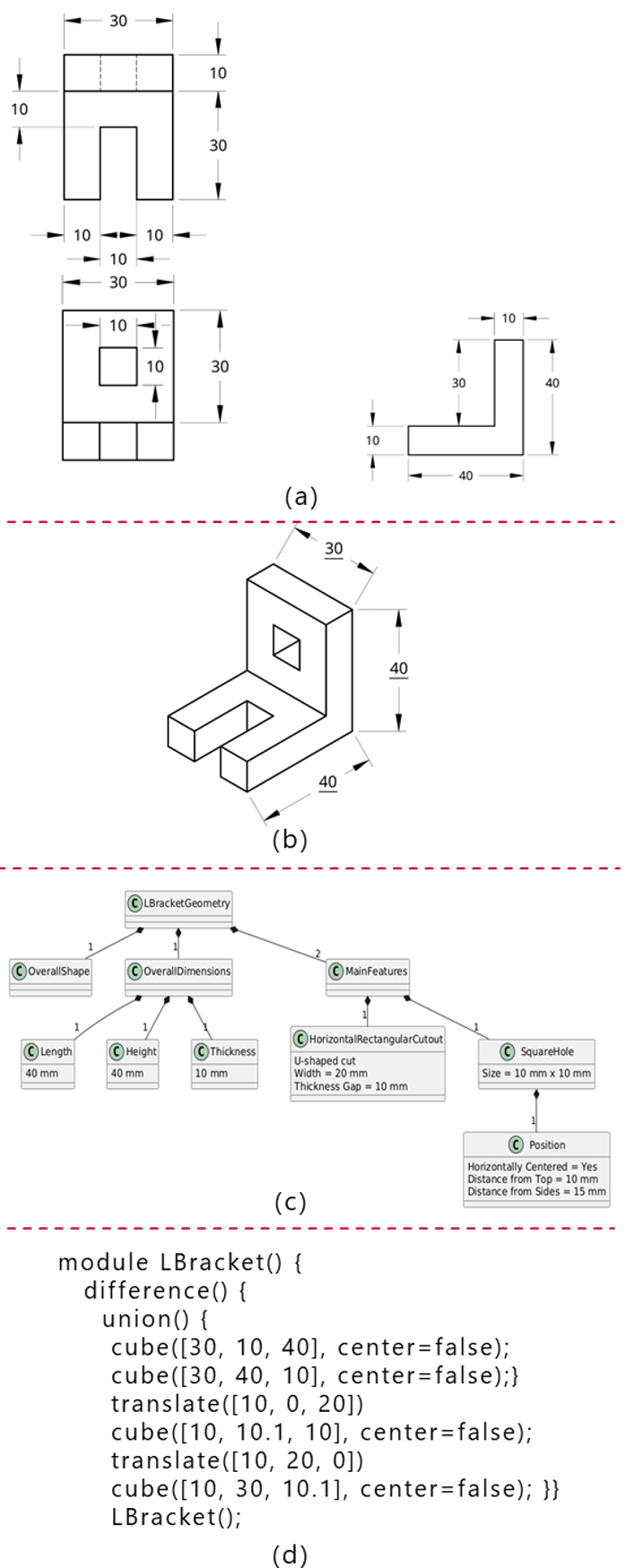}
    \caption{Four input modalities for the L-bracket design: 
    (a) Three-view orthographic drawings, 
    (b) Isometric sketch, 
    (c) Geometric structure representation, 
    (d) Code-level correction prompt.}
    \label{fig:LLMinput}
\end{figure}

\vspace{0.4cm}

Figure~\ref{fig:LLMinput} illustrates four input modalities employed to prompt the LLM, each progressively structured to convey increasing semantic richness and structural clarity:

\begin{itemize}
    \item \textbf{Three-view orthographic drawings:} Standard 2D engineering projections (top, front, side) including basic dimensions (Figure~\ref{fig:LLMinput}a). These views provide foundational geometric information but inherently lack explicit depth cues, feature hierarchy, and spatial context.

    \item \textbf{Isometric sketch:} A 3D representation offering visual spatial relationships and implicit depth information (Figure~\ref{fig:LLMinput}b). This modality enhances the LLM's ability to infer accurate surface and volumetric properties, reducing ambiguity present in purely orthographic representations.

    \item \textbf{Geometric structure representation:} A structured semantic tree explicitly detailing the part's parametric attributes and logical component hierarchy (e.g., "L-shaped base with 10 mm thick arms and a 10×10 mm hole") (Figure~\ref{fig:LLMinput}c). This structured textual description significantly aids the LLM in understanding precise geometric and functional characteristics.

    \item \textbf{Code-level correction prompt:} Explicit natural language instructions designed to complete or correct a partially written OpenSCAD script (Figure~\ref{fig:LLMinput}d). This highly structured symbolic modality leverages the LLM's reasoning capabilities over formal geometry representations, ensuring maximal accuracy and interpretability.
\end{itemize}

The method used to generate and evaluate 3D shapes from these inputs is based on the iterative pipeline depicted in Figure~\ref{fig:system_architecture}. During this experiment, we used OpenAI GPT-4.5 to generate openSCAD code ~\citep{sadik2023coding}. For the initial three modalities, we provide the LLM with a straightforward instruction such as "use this input to construct OpenSCAD code." The resulting OpenSCAD code is visualized using the OpenSCAD library to generate corresponding STL models. To quantitatively evaluate the outputs, we employ a Python script comparing the generated STL models against the ground-truth STL using the previously defined structural complexity and geometric similarity metrics. Iterations continue based on quantitative feedback until no further significant improvements are observed. At that point, the human designer decides to move on to the next input modality, recognizing that better results are unlikely with the current one—especially if the quantitative evaluations begin to diverge or plateau. For code-level correction prompts, errors identified through quantitative evaluation are addressed directly through manual code adjustments, without further involvement of the LLM.

%%%%%%%%%%%%%%%%%%%%%%%%%%%%%%%%%%%%%%%%%%%%%%%%%%%%%%%%%%%%%%%%%%%%%%%%

\subsection{Generated 3D Models Evaluation}

The generated output highlights how the proposed human-in-the-loop architecture effectively enhances the generation fidelity of LLM-created 3D models across progressively informative input modalities. As illustrated in Figure~\ref{fig:EvalLines}, there is a clear upward trend across all similarity metrics as inputs evolve from three-view drawings to isometric sketches, geometric structure descriptions, and finally code-level correction prompts. To ground the evaluation, we first computed the structural complexity of the ground-truth L-bracket model using the composite metric described in Section~\ref{sec:evaluation}. The final complexity score was calculated to be $C = 255$, indicating a moderately complex model with significant topological and geometric detail. This value serves as a baseline for interpreting the fidelity and difficulty of the generation task across different input modalities.

\begin{figure}[!ht]
    \centering
    \includegraphics[width=0.9\linewidth]{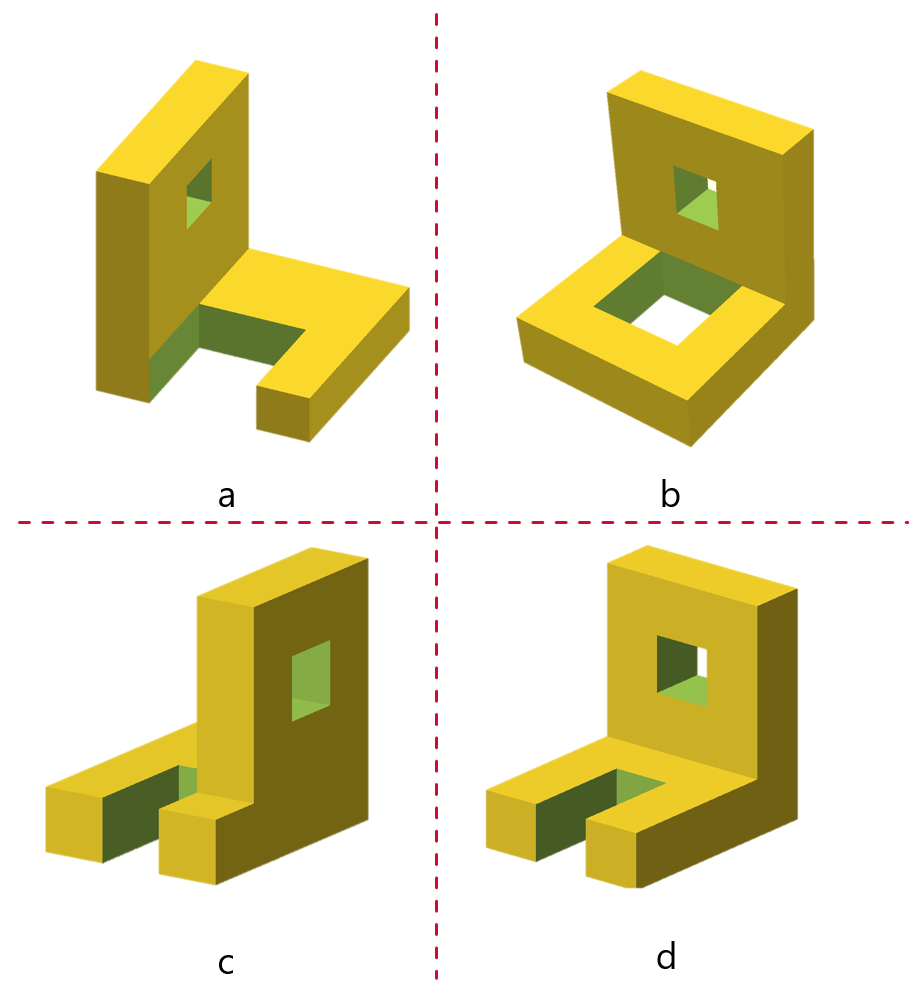}
    \caption{Generated 3D models: 
    (a) from three-view input, 
    (b) from isometric input, 
    (c) from geometric structure input, 
    (d) from code-based correction.} \vspace{2em} 
    \label{fig:LLMoutput}
\end{figure}

\begin{figure}[!htbp]
    \centering
    \includegraphics[width=\linewidth]{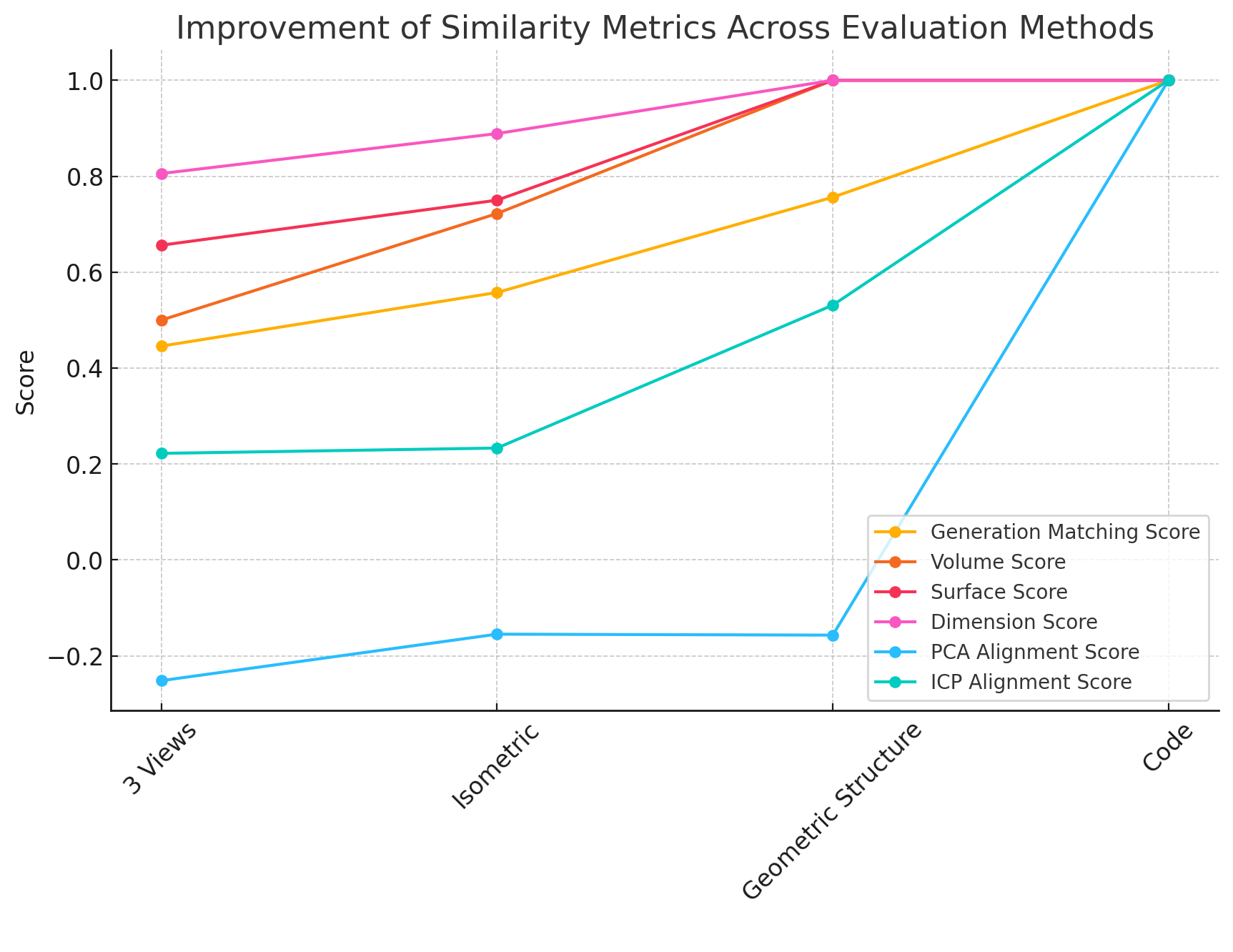}
    \caption{Improvement of similarity metrics across input methods. Higher values indicate better shape fidelity; lower Hausdorff distance indicates reduced maximum error.} \vspace{2em} 
    \label{fig:EvalLines}
\end{figure}

Initially, 3-view orthographic inputs resulted in substantial errors, particularly in spatial orientation and feature misplacement. However, through the iterative human-in-the-loop process—consisting of prompt refinement, output visualization, and quantitative feedback—the system progressively improved model fidelity. The isometric modality demonstrated moderate improvements, helping the LLM better infer depth and shape relationships. When using geometric structure trees, the LLM produced well-proportioned and semantically accurate models, achieving high scores across volume, surface, and dimensional metrics. An interesting observation arises in the geometric structure case (Figure~\ref{fig:LLMoutput}c). The generated model exactly matched the ground truth in terms of volume, dimension, and surface area, as validated by the quantitative evaluation. However, the top arm of the bracket was rotated, which led to misalignment indicators such as PCA and ICP scores deviating from ideal values. These rotation and displacement metrics were critical in guiding the human reviewer to pinpoint the exact discrepancy. The insight enabled minimal intervention—just a few parameter modifications in the OpenSCAD code—to correct the orientation and achieve complete model fidelity.

Crucially, the code-level correction prompt, supported by precise symbolic descriptions, achieved perfect or near-perfect scores across all metrics, including PCA and ICP alignment and zero Hausdorff distance. These results confirm that symbolic inputs enable the LLM to interpret and reproduce the intended geometry with the highest precision. The iterative improvement enabled by the human-in-the-loop pipeline is central to this progression. By incorporating human feedback at each evaluation stage, errors were systematically identified and rectified, reinforcing the LLM’s ability to correct its outputs based on quantifiable feedback. This feedback loop not only improves final output quality but also reduces iteration cycles in later stages.

Figure~\ref{fig:LLMoutput} presents the LLM-generated 3D models corresponding to each input type. A clear progression in model accuracy and completeness is observed as the input becomes more semantically informative. The model generated from 3-view drawings (a) exhibits significant misinterpretations, including orientation and cutout misplacements. The isometric input (b) improves depth inference and hole positioning. The geometric structure input (c) enables a more complete shape with accurate proportions, while the code-based prompt (d) results in a near-perfect reconstruction, showcasing precise geometry and clean topology.

\begin{table}[!htbp]
\centering
\caption{Evaluation results for generated models.}
\resizebox{\columnwidth}{!}{%
\begin{tabular}{lccccccc}
\toprule
\makecell{\textbf{Input}\\\textbf{Type}} & \makecell{\textbf{Gen.}\\\textbf{Score}} & \makecell{\textbf{Vol.}\\\textbf{Score}} & \makecell{\textbf{Surf.}\\\textbf{Score}} & \makecell{\textbf{Dim.}\\\textbf{Score}} & \makecell{\textbf{PCA Align.}\\\textbf{Score}} & \makecell{\textbf{ICP Align.}\\\textbf{Score}} & \makecell{\textbf{Hausdorff}\\\textbf{Dist.}} \\
\midrule
3 Views             & 0.4458 & 0.5000 & 0.6562 & 0.8056 & -0.2516 & 0.2222 & 33.1662 \\
Isometric           & 0.5576 & 0.7222 & 0.7500 & 0.8889 & -0.1547 & 0.2333 & 14.1421 \\
\makecell{Geometric \\ Structure} & 0.7562 & 1.0000 & 1.0000 & 1.0000 & -0.1567 & 0.5312 & 22.3607 \\
Code                & \textbf{1.0000} & \textbf{1.0000} & \textbf{1.0000} & \textbf{1.0000} & \textbf{1.0000} & \textbf{1.0000} & \textbf{0.0000} \\
\bottomrule
\end{tabular}%
}
\label{tab:evaluation_results}
\end{table}

The evaluation results, summarized in Table~\ref{tab:evaluation_results} and visualized in Figure~\ref{fig:EvalLines}, quantitatively reinforce the qualitative improvements observed in the 3D outputs. Across all similarity metrics—volume, surface, dimension, alignment, and Hausdorff distance—there is a clear upward trend as the input becomes more structured.

Notably:
\begin{itemize}
    \item \textbf{3-View drawings} led to the lowest performance across metrics, especially in PCA and ICP alignment scores, indicating poor shape orientation and registration.
    \item \textbf{Isometric input} improved volumetric and surface fidelity but still suffered from geometric ambiguity in less-defined features.
    \item \textbf{Geometric structure representation} significantly enhanced all metrics due to explicit shape semantics and hierarchy.
    \item \textbf{Code-level correction} achieved perfect scores across all metrics, highlighting the advantage of symbolic input for precision and interpretability.
\end{itemize}

These findings suggest that LLMs can reconstruct high-fidelity 3D models from diverse input types, but their performance depends heavily on the clarity and structure of the input. While visual sketches offer a starting point, combining them with semantic representations or symbolic CAD code unlocks far greater accuracy. This underscores the importance of input modality design in AI-assisted CAD workflows.

%%%%%%%%%%%%%%%%%%%%%%%%%%%%%%%%%%%%%%%%%%%%%%%%%%%%%%%%%%%%%%%%%%%%%%%%
\section{Discussion, Conclusion, and Future Work}
\label{sec:conclusion}

This work has presented a comprehensive evaluation framework for assessing 3D models generated by LLMs, focusing on how input modality and human-in-the-loop iteration influence the geometric and structural fidelity of outputs. Our methodology, grounded in a rigorous suite of structural complexity and geometric similarity metrics, allowed us to systematically benchmark model fidelity across varying degrees of semantic richness in inputs—from conventional 2D drawings to symbolic CAD code prompts.

The core insight from our case study on an L-bracket component is that input expressiveness plays a crucial role in the accuracy of LLM-generated models. 3-view orthographic inputs, despite being a staple in engineering communication, yielded limited performance due to their lack of depth and structural hierarchy. Isometric sketches improved spatial interpretation, while geometric structure trees introduced semantic clarity that substantially boosted volumetric and dimensional accuracy. Notably, in the geometric structure case, the LLM-generated model achieved an exact match with the ground truth in terms of volume, surface area, and key dimensions. Misalignment was restricted to a single rotation of the upper bracket, as confirmed by PCA and ICP alignment metrics—evidence of the importance of such local feedback for final tuning. The code-level correction prompt proved most effective, achieving perfect scores across all metrics. This highlights the strength of symbolic input in enabling the LLM to reason over geometry with high fidelity.

One of the major contributions of this study is the demonstration that a quantitative evaluation strategy enables significantly faster convergence to the ground truth geometry compared to traditional, visually guided methods. Instead of relying on qualitative assessment and human intuition alone, our approach offers precise and measurable feedback that facilitates targeted corrections, thereby streamlining the iteration process. This finding has practical implications for improving efficiency and accuracy in AI-assisted CAD workflows.

Moreover, the structural complexity score of the target model ($C = 255$) established a realistic benchmark that LLMs were ultimately able to meet with appropriate input structuring and iteration. The complexity metric not only guided the interpretation of shape fidelity but also served as a diagnostic tool for understanding generation limitations across input modalities. Importantly, our evaluation framework is designed to be both modular and agnostic to the generation method or architecture. This flexibility allows it to be applied in benchmarking and comparing outputs from diverse LLMs, generation pipelines, or hybrid models—making it a versatile tool for researchers working across different methodologies in 3D generation.

Looking ahead, several directions merit further exploration. First, the structural complexity score has shown promise as a valuable diagnostic signal to understand the limitations of different input modalities and generation strategies. This suggests its potential utility in large-scale studies involving diverse datasets to map model capabilities and failure modes systematically. Second, the current formulation of complexity and similarity scores relies on fixed coefficients; future research could investigate dynamic or learned weightings that better capture task-specific importance or user-defined priorities. Additionally, this framework should be extended to multi-part assemblies and parametrically defined geometries. Hybrid prompting strategies that combine sketches with semantic descriptions or code fragments may reduce ambiguity in early-stage inputs. Finally, tighter integration between human evaluators and generation agents—perhaps via real-time feedback or interactive correction tools—could further streamline the generation-refinement loop.

In conclusion, our study confirms the LLMs potential in high-precision 3D generation workflows and provides a structured path forward for improving performance and usability in CAD-based applications through measured human-LLM synergy. Our proposed evaluation framework contributes a reusable, model-agnostic methodology that advances rigorous benchmarking in this emerging research area.

%%%%%%%%%%%%%%%%%%%%%%%%%%%%%%%%%%%%%%%%%%%%%%%%%%%%%%%%%%%%%%%%%%%%%%%%
\section*{Appendix: OpenSCAD Code for Figure 3 Models}

The following OpenSCAD code listings correspond to the generated 3D models shown in Figure~\ref{fig:LLMoutput}.

\begin{lstlisting}[language=c++,caption={OpenSCAD model for Figure 3a: Three-view input},label={lst:model_a}]
module L_Bracket_a() {
    difference() {
        union() {
            cube([40, 40, 10]); // Base
            translate([0, 0, 10])
                cube([10, 40, 40]); // Vertical plate
        } // U-shaped cutout in the base
        translate([10, 0, 0])
            cube([20, 20, 10.1]); // Slightly oversized in Z
        // Square hole in the vertical plate
        translate([0, 15, 30])
            cube([10.1, 10, 10]); // Oversized in X for clean subtraction
    }
}
L_Bracket_a();
\end{lstlisting}

\begin{lstlisting}[language=c++,caption={OpenSCAD model for Figure 3b: Isometric input},label={lst:model_b}]
module L_Bracket_b() {
    difference() {
        union() {
            cube([40, 10, 40]); // Vertical leg
            cube([40, 40, 10]); // Base plate
        }
        // Square hole in the vertical leg
        translate([15, 0, 20])
            cube([10, 10.1, 10]); // Oversized in Y
        // U-shaped cutout in the base
        translate([10, 10, 0])
            cube([20, 20, 10.1]); // Oversized in Z
    }
}
L_Bracket_b();
\end{lstlisting}

\begin{lstlisting}[language=c++,caption={OpenSCAD model for Figure 3c: Geometric structure input},label={lst:model_c}]
module L_Bracket_c() {
    difference() {
        union() {
            cube([10, 30, 40]); // Vertical leg
            cube([30, 40, 10]); // Base plate
        }
        // Horizontal rectangular cutout
        translate([10, 20, 0])
            cube([10, 30, 10.1]); // Slightly oversized
        // Vertical square hole
        translate([0, 10, 20])
            cube([10.1, 10, 10]); // Clean subtraction
    }
}
L_Bracket_c();
\end{lstlisting}

\begin{lstlisting}[language=c++,caption={OpenSCAD model for Figure 3d: Code-based correction},label={lst:model_d}]
module L_Bracket_d() {
    difference() {
        union() {
            cube([30, 10, 40]); // Vertical leg
            cube([30, 40, 10]); // Base plate
        }
        // Square hole in the vertical leg
        translate([10, 0, 20])
            cube([10, 10.1, 10]); // Oversized in Y
        // U-shaped cutout in the base
        translate([10, 20, 0])
            cube([10, 30, 10.1]); // Oversized in Z
    }
}
L_Bracket_d();
\end{lstlisting}
%%%%%%%%%%%%%%%%%%%%%%%%%%%%%%%%%%%%%%%%%%%%%%%%%%%%%%%%%%%%%%%%%%%%%%%%

%%% Use this command to include your bibliography file.

\bibliography{main}

\end{document}